\documentclass{article}
\usepackage{amsmath,amsfonts,amssymb,graphicx,mlspconf}
\usepackage{mathrsfs}
\usepackage{color}

\copyrightnotice{978-1-5386-5477-4/18/\$31.00 {\copyright}2018 IEEE}

\toappear{2018 IEEE International Workshop on Machine Learning for Signal Processing, Sept.\ 17--20, 2018, Aalborg, Denmark}

\def\vvec#1{\mathsf{vec}\left(#1\right)}

\def\RR{{\mathbb R}}

\def\EE{{\mathbb E}}

\def\Ex#1{{\EE\left\{#1\right\}}}

\def\SNR{\mathsf{SNR}}

\def\Tr#1{{\mathsf{Tr}\!\left\{#1\right\}}}

\def\be{\begin{equation}}
\def\beq#1{\begin{equation}\label{#1}}
\def\ee{\end{equation}}
\def\bea{\begin{eqnarray}}
\def\beqa#1{\begin{eqnarray}\label{#1}}
\def\eea{\end{eqnarray}}
\def\ba{\begin{array}}
\def\ea{\end{array}}

\DeclareMathAlphabet{\mathpzc}{OT1}{pzc}{m}{it}

\def\cF{{\mathcal F}}

\def\cN{{\mathcal N}}

\def\ccD{{\mathscr{D}}}

\def\blambda{\boldsymbol{\lambda}}
\def\bomega{\boldsymbol{\omega}}

\def\ba{{\mathbf a}}

\def\bd{{\mathbf d}}

\def\bn{{\mathbf n}}

\def\bO{{\mathbf 0}}

\def\bw{{\mathbf w}}

\title{Space-time extension of the MEM approach\\ for electromagnetic neuroimaging}
\name{M.C. Roubaud$^1$, J.M. Lina$^{2,3}$, J. Carrier$^3$ and B. Torr\'esani$^1$}
\address{$^1$ Aix Marseille Univ, CNRS, Centrale Marseille, I2M, Marseille, 
France\\
$^2$ Department of Electrical Engineering, Ecole de Technologie Sup\'erieure, Montr\'eal, Qu\'ebec, Canada\\
$^3$ Centre d'\'etudes avanc\'ees en m\'edecine du sommeil, Universit\'e de Montr\'eal, Qu\'ebec, Canada}
\begin{document}
%

\maketitle
\begin{abstract}
The wavelet Maximum Entropy on the Mean (wMEM) approach to the MEG inverse problem is revisited and extended to infer brain activity from full space-time data. The resulting dimensionality increase is tackled using a collection of techniques, that includes time and space dimension reduction (using respectively wavelet and spatial filter based reductions), Kronecker product modeling for covariance matrices, and numerical manipulation of the free energy directly in matrix form. This leads to a smooth numerical optimization problem of reasonable dimension, solved using standard approaches.

The method is applied to the MEG inverse problem. Results of a simulation study in the context of slow wave localization from sleep MEG data are presented and discussed.
\end{abstract}

\section{Introduction}
\label{sec:intro}
EEG and MEG inverse problems are notoriously difficult ill posed inverse problems. The corresponding electric potentials or magnetic fields measurements are far from sufficient to yield a unique solution, which can only be obtained by adding constraints or regularization.
Most classical approaches (MNE, wMNE, LORETA, MCE, see~\cite{Baillet13forward} and references therein) do not explicitly use time dependence: source parameters are estimated at each time point and time correlations are not  directly exploited (unless some pre-processing techniques used prior to inverse problem resolution).
All those solutions can be interpreted as maximum \textit{a posteriori} estimates, with gaussian noise distribution and various prior choices (often Gaussian or Laplacian).

An alternative approach was proposed in~\cite{Lina14wavelet}, based on the Maximum Entropy on the Mean (MEM) principle combined with wavelet representation of time courses. This Bayesian technique introduces more freedom in the modeling, and yields smooth optimization problems of dimension much smaller  than the number of sources. Nevertheless, while time correlations are to some extent captured by wavelet coefficients, inversion is still performed coefficientwise.

We describe here a space-time extension of the wMEM approach of~\cite{Lina14wavelet} in which time dependence is explicitely modeled and accounted for. To overcome the curse of dimensionality, we rely on three main ingredients: the Kronecker product factorization of covariance matrices (the noise covariance, in the spirit of~\cite{Bijma03mathematical}, and the source covariances), the matrix formulation of the MEM optimization problem, which generates considerable savings, and space and time dimension reduction techniques. The resulting (smooth and concave) optimization problem is solved using standard tools.

We apply this approach to the MEG inverse problem, on a dataset originating from a study of slow waves in deep sleep MEG signals~\cite{Lina16electromagnetic}. Results on real data will be discussed in a forthcoming publication, we present here results of a simulation study that shows the ability of the approach to localize sources with such a given time course.

The paper is organized as follows. We describe in Sections~\ref{sec:InvProb} and~\ref{sec:matrix} the main aspects of our approach, and provide numerical results in Section~\ref{sec:results}. Section~\ref{sec:conclusion} is devoted to conclusions and discussion.

\section{Problem statement, MEM approach}
\label{sec:InvProb}
\subsection{Problem statement, notations}
Observed signals are modeled as multivariate time series $\ell\to z(\ell)\in\RR^{J_0}$ ($J_0$ being the number of sensors), and similarly the observation noise writes $\ell\to n_0(\ell)\in\RR^{J_0}$. In the framework of the distributed sources model, we denote by $K$ the number of mesh points on the cortex, and by $\ell\to x(\ell)\in\RR^K$ the corresponding time courses. Denoting by $G_0\in\RR^{J_0\times K}$ the lead-field matrix that summarizes the forward problem (propagation from the cortical surface to the sensors), this leads to the time domain observation equation
\begin{equation}
z(\ell) = G_0 x(\ell) + n_0(\ell)\ ,\quad \ell=1,\dots L_0\ ,
\end{equation}
$L_0$ being the number of time samples. These multivariate time series are reshaped as matrices $Z,N_0\in\RR^{L_0\times J_0}$, $X\in\RR^{L_0\times K}$.  This yields the matrix formulation:
\begin{equation}
Z = XG_0^T  + N_0\ ,
\end{equation}
where $z=\mathsf{vec}(Z^T)$. Here ${\cdot}^T$ denotes matrix transposition and $\mathsf{vec}(\cdot)$  vertical concatenation of columns.
After dimension reduction (wavelet transform and projection onto a $L$ dimensional wavelet subspace in time domain, and space domain reduction to $J$ spatial filters, see Section~\ref{se:red} below), we obtain in matrix form
\be
D = W G^T + N = Y H^T\ .
\ee
Here, $D,N\in\RR^{L\times J}$, $W\in\RR^{L\times K}$, and $Y = [W,N]$ (resp. $H=[G;I_J]$) denotes the horizontal (resp. vertical) concatenation of matrices $W$ and $N$ (resp. $G$ and $I_J$).  This is the inverse problem that will be of interest to us here.

\subsection{The vector MEM approach}

The Maximum Entropy on the Mean (MEM, see~\cite{gamboa1997,LeBesnerais99new} for presentations) is a generic Bayesian approach for solving linear inverse problem. It was first used for the MEG inverse problem in~\cite{Lina14wavelet}, combined with wavelet decomposition, resulting in the so-called wMEM method from which this work is strongly inspired. Let $d$ be a column of $D$ and $y$ the corresponding column of $Y$. MEM models the ``source and noise" vector $d$ as a random vector, and the observation equation $d=Hy$ is replaced with $\Ex{Hy}=d$, $\Ex{\cdot}$ denoting expectation. MEM requires specifying a reference probability distribution on $y$. Noise and sources are assumed independent, and MEM seeks the probability distribution on $y$ that minimizes the Kullback-Leibler divergence to the reference distribution, under the constraint $\Ex{Hy}=d$. Sources are finally estimated as the expectation $\Ex{y}$ of $y$ with respect to the so-obtained distribution.

This constrained minimization problem turns out~\cite{gamboa1997,LeBesnerais99new} to be equivalent to the maximization of a concave function $\ccD(\lambda)$ of an auxiliary variable $\lambda$, whose dimension equals the dimension of observations $z$,
\be
\label{fo:MEM.free.energy}
\ccD(\lambda) = \lambda\cdot d -\cF_y^*(H^T\lambda)\ ,
\ee
where $\lambda\cdot d$ is the inner product in the observation space, and the function $\cF_y^*$, called log-partition function, is fully specified by the reference probability distribution. Given the optimizer $\lambda_*= \mathsf{arg\,min}_\lambda\ccD(\lambda)$, the estimate $\hat y$ for the ``sources and noise" vector finally reads
\be
\label{fo:MEM.sol}
\hat y = H\nabla \ccD(\lambda_*)\ .
\ee

In~\cite{Lina14wavelet}, this approach was proposed and tested for the MEG inverse problem. In this approach, the multi-sensor observations $d$ are wavelet coefficients, leading to a $J_0$-dimensional optimization problem (which is a major asset of the approach, given that $J_0\ll K$). The reference model is a Gaussian mixture model, based upon a parcellization of the cortex into independent regions. An important aspect is the parametrization of the reference model, for which the authors propose a generic strategy exploiting the cortex geometry and the source pre-localization MSP technique of~\cite{Mattout05Multivariate}.

The extension to (vectorized) space-time data is straightforward, but increases significantly the dimension of the optimization problem, which in addition involves calculations in a very high dimensional space (cortex $\times$ time). The latter can be simplified by suitable prior choices (described in Section~\ref{sec:matrix}) and matrix formulation. However, dimension reduction in the (space-time) observation space is also necessary. Notice that the space-time extension also increases the number of parameters in the model.

\subsection{Dimension reduction}
\label{se:red}
Wavelet transform~\cite{Mallat08wavelet} provides alternative representations for signals, that often have the property of concentrating the relevant information in a small number of coefficients, and/or enforcing decorrelation, yielding sparse or diagonal dominant covariance matrices. We perform a channel-wise orthonormal wavelet transform, followed by a selection of the relevant coefficients. Here the retained coefficients are selected \textit{a priori}, and are the same for all trials (wavelet coefficients with largets trial averaged energy are selected, coefficients perturbed by boundary effects are not retained). The selection could also be done adaptively, we don't address this question here.
Wavelet transform and coefficient selection result in an observation matrix $W_1\in\RR^{L\times J_0}$, $L$ being the number of retained coefficients, generally $L<J_0$.

The sensor domain can also be reduced using a spatial filtering. Several approaches could be chosen (see e.g.~\cite{Sekihara08adaptive} for a review), we limit ourselves here to simple PCA-based dimension reduction: projection onto the first $J$ principal axes. Among possible extensions, discriminant filters such as introduced in~\cite{Spinnato14finding} are an interesting perspective.

After time and space dimension reduction, the observed data takes the form of a matrix $D\in\RR^{L\times J}$, and the inverse problem to be solved writes
\be
D = W G^T +N \ , 
\ee
where $G$ and $N$ are the projections of the lead-field matrix $G_0$ and the noise $N_0$ on the reduced sensor space. $W\in\RR^{L\times K}$ represents the unknown source wavelet coefficients.

\section{Matrix wMEM for space-time data}
\label{sec:matrix}
We now turn to the time-space model. The data to be processed is a time-space  matrix $D$ of size $L\times J$. We denote corresponding vectorized matrices with boldface lowercase symbols: observations $\bd = \vvec{D^T}$, noise $\bn = \vvec{N^T}$, wavelet source coefficients $\bw = \vvec{W^T}$ and so on.

\subsection{Matrix gaussian distribution, Kronecker covariance factorization}
Our model is based upon the following assumptions and models on noise and cortical sources.
\subsubsection{Noise model}
The observation noise $\bn$ is distributed following a matrix normal law~\cite{gupta99matrix}, i.e.
\be
\bn=\vvec{N^T}\sim\cN(\bO,\Sigma_N^t\otimes\Sigma_N^s)
\ee
with Kronecker covariance matrix (as proposed in~\cite{Bijma03mathematical}) where $\Sigma_N^t\in\RR^{L\times L}$ is the time covariance matrix, and $\Sigma_N^s\in\RR^{J\times J}$ is the space (i.e. sensor) covariance matrix.
Notice that this factorized model requires estimating and storing $L^2 + J^2$ numbers rather than $L^2J^2$.
It is worth mentioning that Kronecker product covariance matrices can be estimated using dedicated algorithms (the so-called flip-flop algorithms), whose convergence has been studied and proven~\cite{Dutilleul99MLE,Srivastava09models}.
\subsubsection{Source model}
Recall that the source space is a mesh of the cortical surface, of size $K$.
The source model is based upon a parcellization of the cortical surface, i.e. a segmentation into $P$ connected regions, called parcels, of size $K_p,\,p=1,\dots P$.

The reference distribution on sources is then written as the product of $P$ independent parcel laws. Following~\cite{Lina14wavelet}, activity in each parcel $\bw_p\in\RR^{LK_p}, p=1,\dots P$ is distributed according to a mixture of two matrix normal distributions, with respective means $\bomega_p$ and $\bO$, and covariance matrices $\Sigma_p = \Sigma_p^t\otimes\Sigma_p^s$ ($\Sigma_p^t\in\RR^{L\times L}$, $\Sigma_p^s\in\RR^{K_p\times K_p}$) and $v_p I_{LK_p}$:
\be
\bw_p\sim\alpha_p\cN(\bomega_p,\Sigma_p)+ (1-\alpha_p)\cN(\bO,v_p I_{LK_p})\ .
\ee
State 1 is the ``active state" (non-zero mean), and state 0 is the ``silent state" (zero mean white noise), coefficient $\alpha_p$ represents the probability for parcel $p$ to be active.

The parcel spatial covariance matrices $\Sigma_p^s$ encode for correlations on the cortex. In the spirit of~\cite{Lina14wavelet}, $\Sigma_p^s$ is set to the restriction to parcel $p$ of the covariance matrix $\exp(-\rho \Delta)$ of a diffusion process on the mesh graph, $\Delta=D-A$ being the graph Laplacian ($A$ and $D$ are respectively the adjacency and degree matrices of the graph), and $\rho$ some fixed parameter (unlike~\cite{Lina14wavelet} which used a truncated series expansion of the exponential, we prefer to stick here to the full matrix exponential, that yields better conditioned covariance matrices).

\subsection{Kronecker-wMEM approach}
We are now in position to describe our approach and solve the inverse problem associated with the model given in matrix form above. Using the parcellized Gaussian mixture model given above, the MEM principle leads to maximize the objective function $\blambda\in\RR^{LJ}\to \ccD(\blambda)$ defined in~\eqref{fo:MEM.free.energy}.

As in~\cite{Lina14wavelet}, the independence of sources and noise, together with the independence of parcels, yield a splitting of the log partition function $\cF^* = \sum_{p=1}^P \cF_p^* + \cF_n^*$.
The assumptions on the noise yield closed form expressions for the objective function $\ccD(\blambda)$ as well as the parcel log-partition functions.
However, these expressions are not easily amenable to numerical calculations as they involve algebraic manipulations in high dimensional space: each $\bw_p$ belongs to a high dimensional space of dimension or the order of $5000$, corresponding matrix vector products are costly. Nevertheless, they can be conveniently re-formulated in matrix form. Defining $\Lambda\in\RR^{L\times J}$ by $\blambda=\vvec{\Lambda^T}$, the objective function writes
\be
\ccD(\Lambda)\! =\! \Tr{D^T\!\Lambda}\! -\! \frac1{2}\Tr{\Lambda^T\Sigma_N^t\Lambda\Sigma_N^s} -\!\sum_{p=1}^P\!\cF_p^*(\Lambda G_p)\, ,
\ee
where $\mathsf{Tr}$ denotes the matrix trace, and $G_p$ is the submatrix of $G$ obtained by restricting to parcel $p$. The parcel log-partition functions are as follows: 
\be
\cF_p^* \!=\! \ln\!\left(\alpha_p \exp\left(\cF_{p,1}^*\right)\! +\! (1\!-\!\alpha_p)\exp\left(\cF_{p,0}^*\right)\right)\ ,
\ee
where the Gaussian log-partition functions are given by 
\begin{eqnarray}
\cF_{p,0}^*(U) &=& \frac{v_p}2\,\Tr{U^TU}\\
\cF_{p,1}^*(U) &=&\Tr{U^T\Omega_p} + \frac{1}{2}\Tr{U^T\Sigma_p^tU\Sigma_p^s}\ ,
\end{eqnarray}
for all $U\in\RR^{L\times K_p}$, and $\Omega_p$ is defined by $\bomega_p=\vvec{\Omega_p^T}$.


Finally, denoting by $\Lambda^*$ the unique optimizer of $\ccD$, the estimate $\widehat{W}_p\in\RR^{LK_p}$ for each parcel $p$ is obtained in matrix form, and given as
\be
\widehat{W}_p = \tilde\alpha_p^*\left[\Omega_p + \Sigma_p^t\Lambda^*G_p\Sigma_p^s\right] + (1-\tilde\alpha_p^*)\,v_p \Lambda^* G_p\ ,
\ee
where the updated activity probabilities read
\be
\tilde\alpha_p \!=\! \frac{\alpha_p}{\alpha_p\! +\! (1\!-\!\alpha_p)\exp\left(\cF_{p,0}^*({\Lambda}^* G_p)\! -\! \cF_{p,1}^*(\Lambda^* G_p)\right)}\ .
\ee
From this the vector form is readily computed as
\be
\hat \bw_p=\vvec{\widehat{W}_p^T}\ ,
\ee
and the time courses of the estimated sources are obtained by inverse wavelet transform.

\section{Numerical results}
\label{sec:results}
The algorithm was implemented in the Matlab\textregistered\ computing environment, using the WaveLab package~\cite{Buckheit05about} for wavelet transform. Numerical optimization of the objective function $\ccD(\lambda)$ was performed using the \textit{minFunc} function~\cite{Schmidt05minFunc}, that implements an adaptive step quasi-Newton (BFGS) algorithm. With the above data, computing time for an inversion is around 5 secs on a laptop (intel core i7-3687U CPU, 2.10GHz $\times$ 4, 16 Go RAM).  

\subsection{Data, and model specifications}
The results presented here originate from a study of slow waves in deep sleep MEG data~\cite{Lina16electromagnetic}. The original dataset (1 subject) consists in 180 trials, 4 seconds long, sampled at 50 Hz, recorded at 272 sensors. The trials were epoched and aligned by experts using EEG data. Besides, 109 ``signal free" recordings were used for noise statistics estimation. Notice that noise contains both sensor noise, and non-interest background activity. Noise time and space covariance matrices were estimated using the flip-flop algorithm~\cite{Dutilleul99MLE}.

Time courses were wavelet transformed using Daubechies 6 orthonormal wavelet basis, after zero padding. Out of 256 wavelet coefficients, 62 most relevant coefficients were selected. Wavelet coefficients influenced by spurious boundary effects were not included. As for sensor domain dimension reduction, data were projected onto the 15 principal components with largest principal values, yielding a $62\times 15$ data matrix for each trial. On original data 15 such virtual channels turn out to capture more than 98\% of inertia, however since we use simulated data here the percentage is even larger (more than 99\%).

The cortical surface (originating from MRI measurements) was discretized with 10002 mesh points, and parcellized into 156 connected parcels (of size ranging from 22 to 175 mesh points, the mean and median being around 60 mesh points), based on anatomical neighborhood criteria.

\smallskip
The model parameters were set as follows, some being fixed, some being estimated from data using a simplified reference model (Gaussian instead of Gaussian mixture, equivalent to wMNE).
Following~\cite{Lina14wavelet}, parcel space covariance matrices $\Sigma_p^s$ were constructed using the graph Laplacian of each parcel, with $\rho=.3$. Assuming time decorrelation (i.e. $\Sigma^t_p=v_p I_L$, with $v_p$ initialized to a constant value, used as signal to noise ratio estimate) and zero mean leads to a Gaussian reference model. Using that reference model, a preliminary estimate for sources was obtained. From the latter, active state parcel means $\bomega_p$ and parcel wavelet covariance matrices $\Sigma_p^t$ were then estimated, to serve as parameters for the Gaussian mixture reference model.
Parcel activity probabilities (which control the sparsity of the source estimates) were set to a constant value, $\alpha_p=.25$.

\subsection{A simulation study}
\begin{figure}
\begin{center}
\includegraphics[scale=.5]{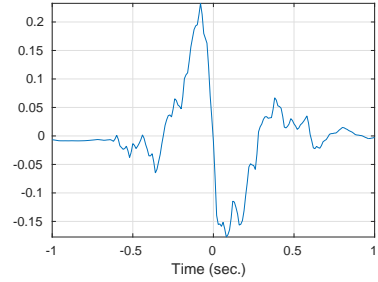}
\end{center}
\vspace{-5mm}
\caption{Time course of the simulated signals}
\label{fi:profil}
\end{figure}

Running the proposed approach on the slow waves dataset turned out to produce a neat time course for the slow wave, which we used in the simulations described below. The time course $\varphi(\ell)$ is displayed in \textbf{Fig.~\ref{fi:profil}}. Simulated datasets were created by 1) generating a connected cortical region $\Omega$ centered at a random seed, 2) generating a current distribution $j_0$ on the cortical surface, set to zero outside $\Omega$ and to $j_0(k,\ell) = \varphi(\ell)$ for $k\in\Omega$, 3) propagating to the sensors using the lead-field matrix, and 4) adding a noise realization taken randomly in the above mentioned ``signal free" trials, with a prescribed signal to noise ratio. We stress that these regions are independent from the parcels used in the inversion algorithm.

From these simulations, performances were evaluated as follows. First a global space-time reconstruction index was computed as the normalized inner product of original cortical sources $j_0(k,\ell)=\varphi(\ell) 1_\Omega(k)$ and reconstructed ones $j_\mathsf{rec}(k,\ell)$:
\[
\iota = \frac{\left\langle j_0,j_\mathsf{rec}\right\rangle}{\|j_0\|\, \|j_\mathsf{rec}\|}\ ,
\]
where inner product and norms are computed with respect to time and space. While the corresponding numerical values are hard to interpret quantitatively, they show a clear improvement when moving from the Gaussian reference to the Gaussian mixture one.
The second criterion rests on decision theory ideas. The reconstructed sources being space-time data, detection of active/silent regions is difficult because activity is time dependent. We focus here on the ability of the method to detect activity with a given time profile, and report on detection performances based upon the following criterion: denoting by $j_\mathsf{rec}(k,\ell)$ the estimated source at time $t$ and on the cortical mesh point $k$, and by $\varphi(\ell)$ the input time profile (the time course displayed in Fig.~\ref{fi:profil}), we compute for each mesh point $k=1,\dots K$ an activity score, denoted by $\kappa$:
\[
\kappa(k) = \frac{\langle j_\mathsf{rec}(k,\cdot),\varphi\rangle}{\|j_\mathsf{rec}(k,\cdot)\|\,\|\varphi\|}\ ,
\]
where inner product and norms are computed with respect to time. This quantity, which ranges from 0 to 1, measures the ability of the method to recover sources with a given time course. For a fixed threshold $\tau$, a cortical mesh point $k$ is declared active if $\kappa(k)\ge \tau$. Varying $\tau$ within $[0,1]$ leads to Receiver Operating Characteristic (ROC) curves, and the area under curve (AUC) is used to assess the performances of the method in terms of detection. The closer AUC to 1 the better the performances, values near 80\% being considered good.

We report here the quantitative evaluations of simulations, for three variants of the method. In the first one (denoted by G), the reference model is a Gaussian model with diagonal source covariance matrix, which provides a closed form solution (equivalent to a variant of the standard wMNE algorithm). The second one (GM) is based upon the Gaussian mixture model described above, whose parameters have been estimated from the Gaussian estimate. In the third one (uGM), the Gaussian mixture MEM algorithm has been run once again, using updated parameters estimated from the latter Gaussian mixture MEM estimate.

Results have been obtained on 100 different source trials (i.e. 100 different connected patches of active sources), averaged over 109 realizations of noise. The results of \textbf{Table~\ref{ta:SNR2}} provide the corresponding average $\iota$-scores and $\kappa$-score based AUCs, which have also been averaged over source trials. The input SNR was set to $\SNR=6\mathsf{dB}$ (i.e. signal was 2 times larger than noise). The table also provides medians and standard deviations. As can be seen, moving from the Gaussian reference model (which does not account for time correlations in the source model) to Gaussian mixture priori significantly improves performances, both in terms of $\iota$⁻score and $\kappa$-based AUCs, with an increase of the standard deviation. Refining the Gaussian mixture reference distribution third column) further sligthtly improves the average and median results, with a further increase in standard deviation.
A closer look at results (not reproduced here) shows that this additional improvement is noticeable when the GM reference is already significantly better than the G rerefence. However, when the quality of reconstruction is not so good (which generally originates from poor quality parameters), updating parameters tends to degrade further the quality of reconstruction.

According to the discussion in~\cite{Lina14wavelet}, AUC results can be biased by the imbalance of silent/active cortical mesh point. For that reason, we also provide corresponding results obtained by (randomly) selecting in each simulation a number of silent cortical mesh points equal to the number of active points. This leads to the so-called restricted AUC (AUC$_R$ in \textbf{Table~\ref{ta:SNR2}}). Differences between AUC and AUC$_R$ are hardly noticeable.

Similar results have been obtained for higher values of input signal to noise ratio (namely, $\SNR=14\mathsf{dB}$ and $\SNR=20\mathsf{dB}$). No striking difference with \textbf{Table~\ref{ta:SNR2}} is to be mentioned, except for the fact that higher SNR improves the $\iota$-index, while AUC means and medians are not really affected.

\begin{table}
\begin{center}
\begin{tabular}{|ll||c|c|c|}
\hline
Criteria&&G&GM&uGM\\
\hline
\hline
&mean&.087&.207&.219\\
\cline{2-5}
$\iota$-index&median&.077&.209&.233\\
\cline{2-5}
&std-dev&.037&.097&.121\\
\hline
\hline
&mean&.710&.833&.839\\
\cline{2-5}
AUC&median&.692&.856&.891\\
\cline{2-5}
&std-dev&.098&.122&.159\\
\hline
\hline
&mean&.710&.833&.839\\
\cline{2-5}
AUC$_R$&median&.694&.857&.890\\
\cline{2-5}
&std-dev&.100&.123&.160\\
\hline
\end{tabular}
\end{center}
\caption{Evaluation for the KMEM reconstruction method for three reference distributions: Gaussian, Gaussian mixture and updated Gaussian mixture. Input SNR was set to 6 dB.}
\label{ta:SNR2}
\end{table}

\subsection{Real data}
\begin{figure}[htb]

\begin{minipage}[b]{.48\linewidth}
  \centering
  \centerline{\includegraphics[width=4cm]{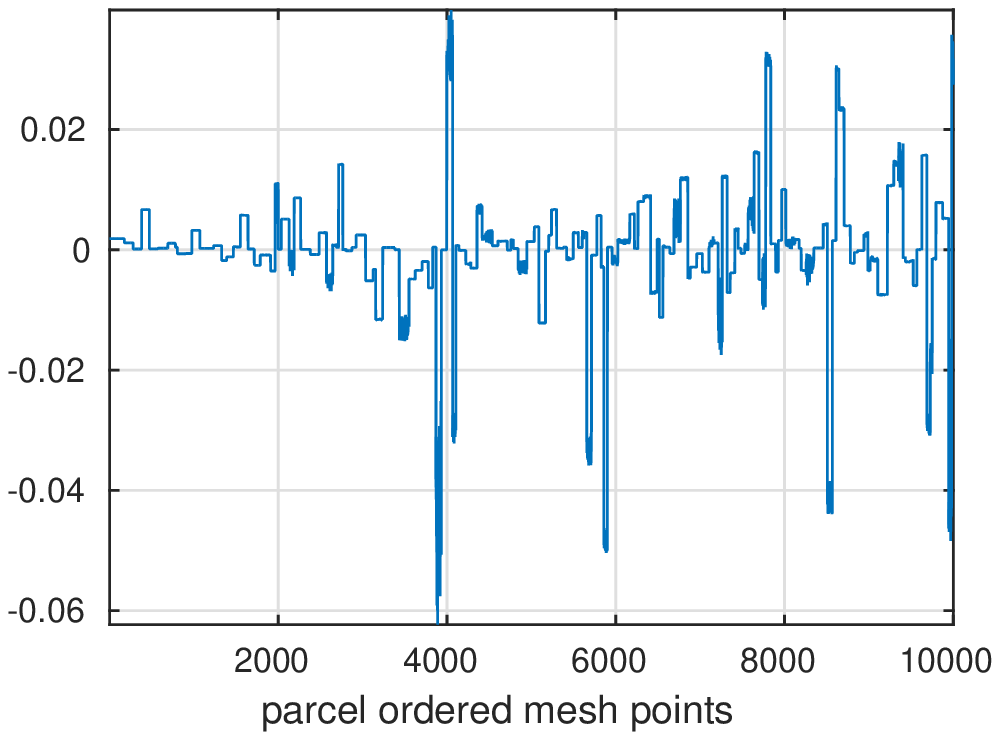}}
  \centerline{(a) PC1: source loadings}\medskip
\end{minipage}
\begin{minipage}[b]{.48\linewidth}
  \centerline{\includegraphics[width=4cm]{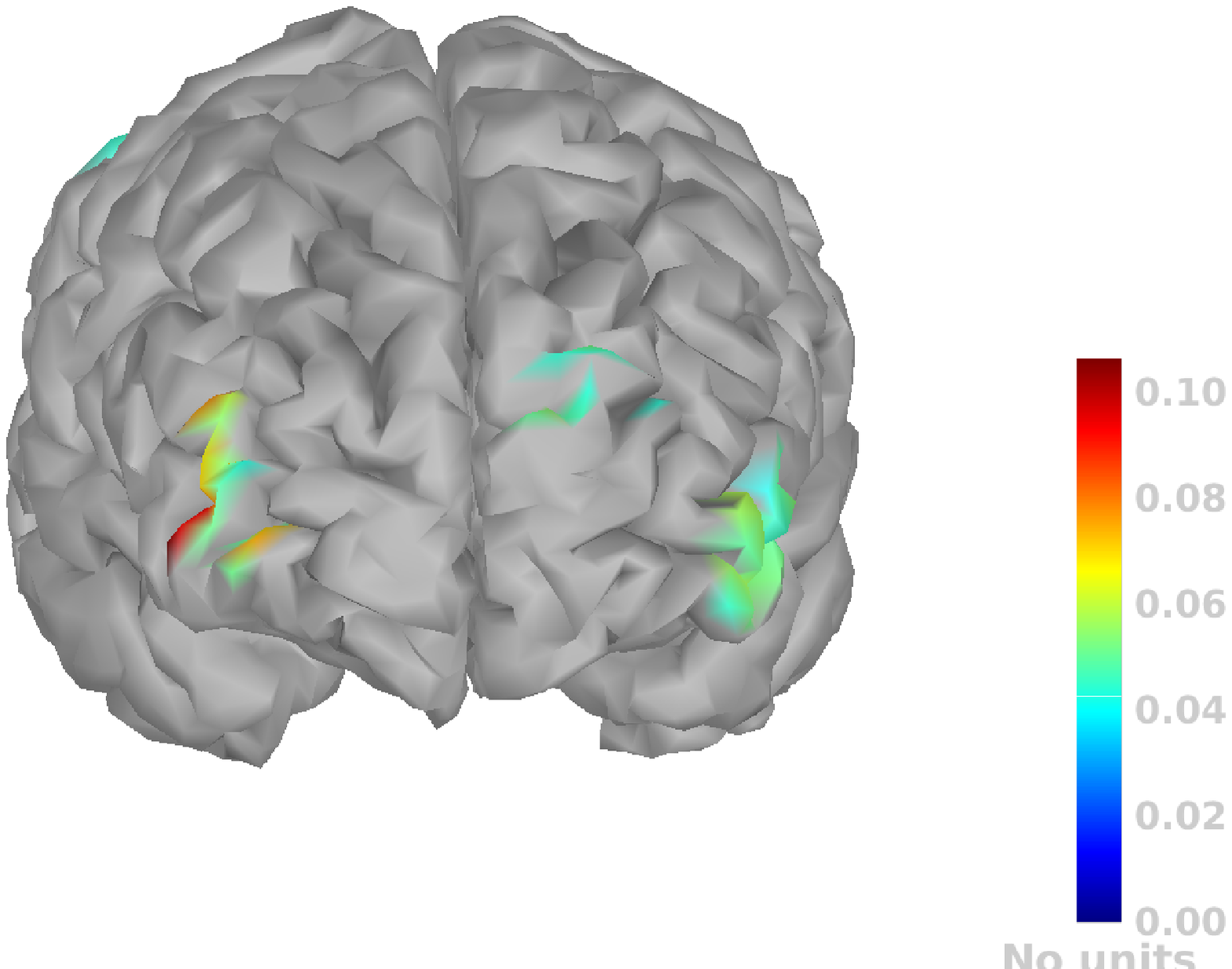}}
  \centerline{(b) PC1 topography: front}\medskip
\end{minipage}
\begin{minipage}[b]{.48\linewidth}
  \centering
  \centerline{\includegraphics[width=4.0cm]{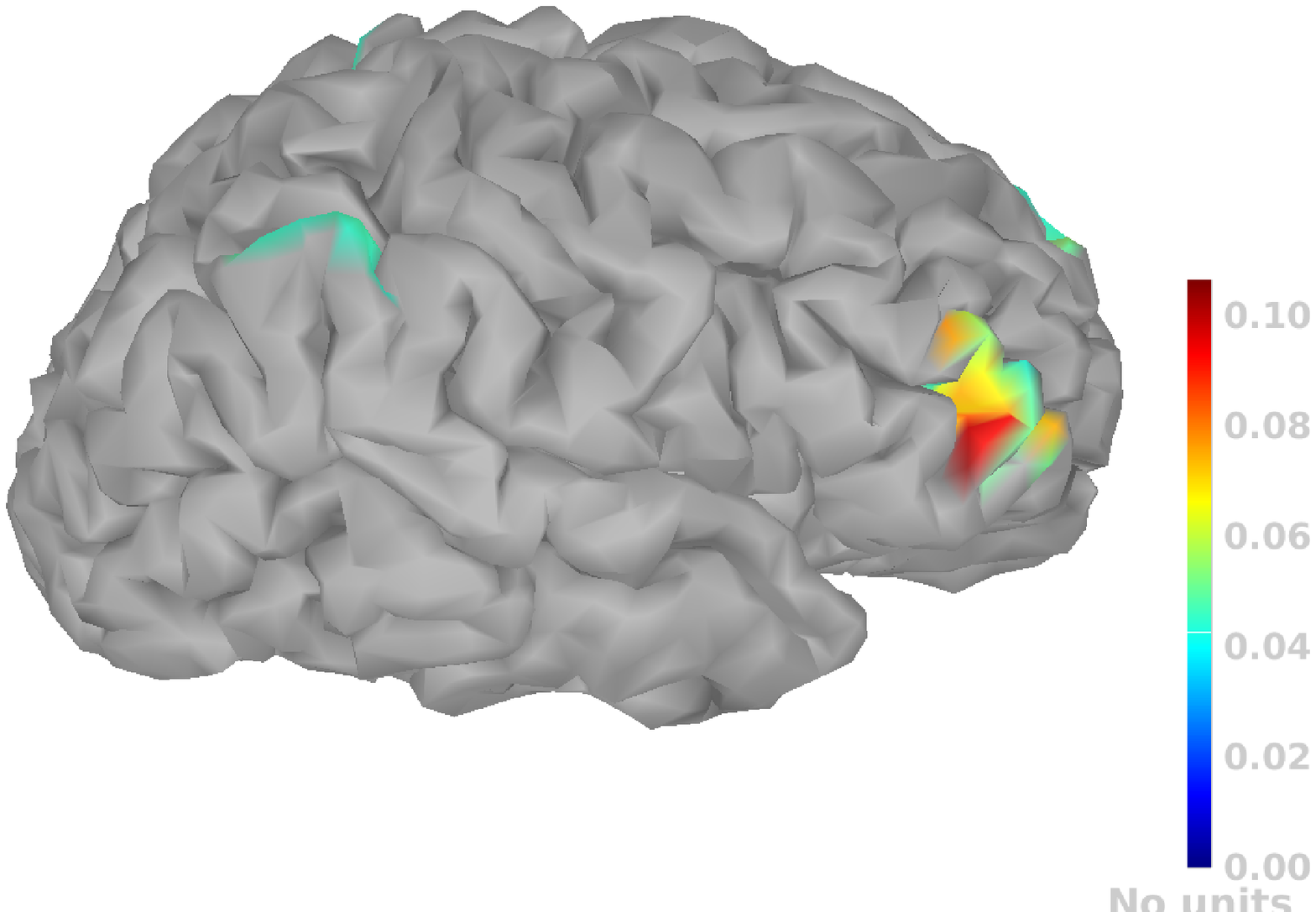}}
  \centerline{(c) PC1 topography: right}\medskip
\end{minipage}
\hfill
\begin{minipage}[b]{0.48\linewidth}
  \centering
  \centerline{\includegraphics[width=4.0cm]{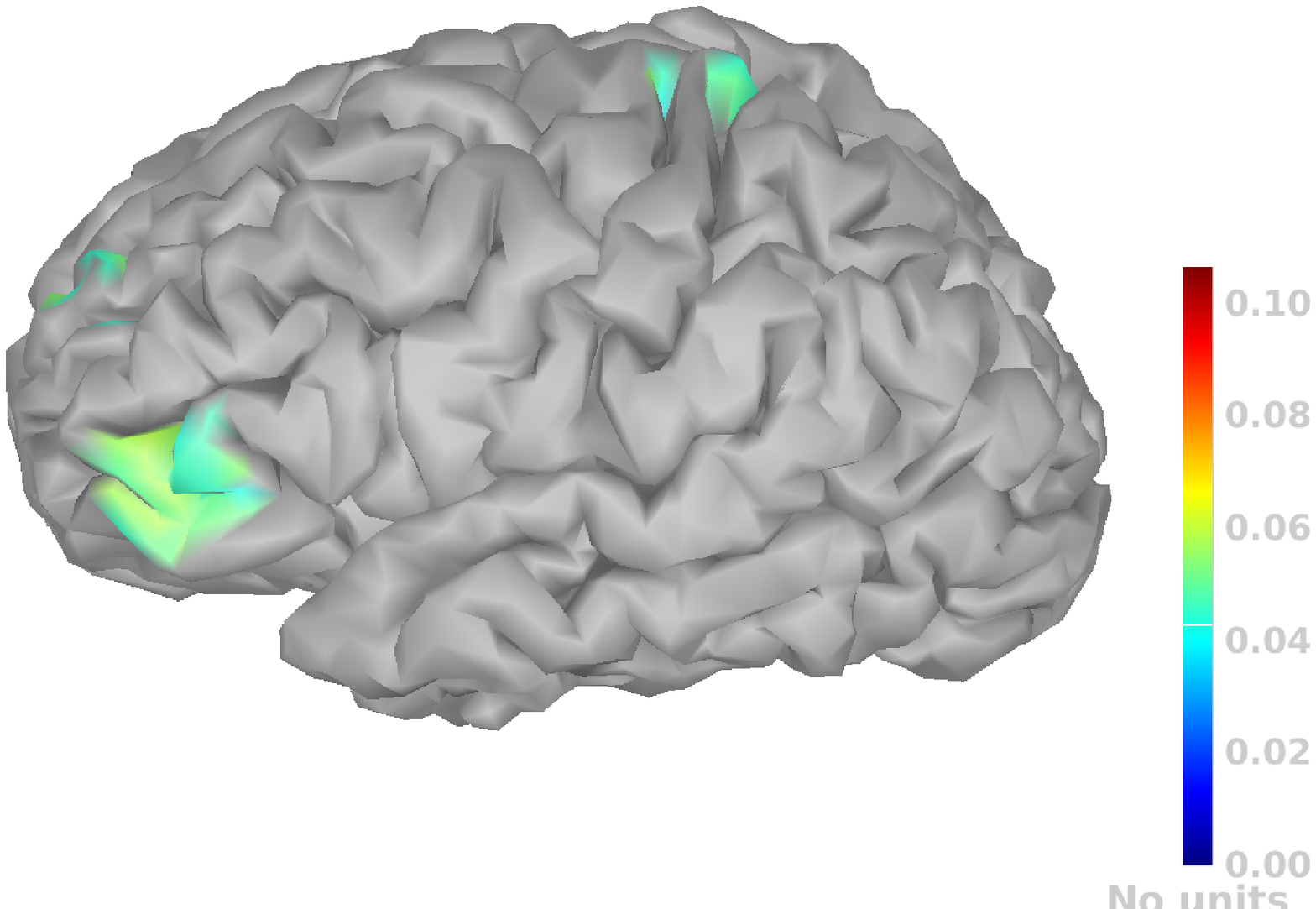}}
  \centerline{(d) PC1 topography: left}\medskip
\end{minipage}
\caption{Spatial projection of the first principal component of estimated sources. a): loadings as a function of mesh points (ordered by increasing parcel number); b,c,d) projection of PC1 loadings onto the cortical surface (threshold set to 40\% of maximal value).}
\label{fi:sources}
\end{figure}

While the present paper focuses on simulation results, the approach has also been tested on real sleep slow waves data. Results will be described and discussed in details in a forthcoming publication, we simply give a short example here. The complete dataset consists in 180 epochs (trials) recorded at $J_0=272$ sensors, which have been co-registered by an expert using additional EEG recordings, to be centered on a slow wave. Sampling rate was 50Hz, the duration of each epoch being 4secs. We report here on results obtained on the trial average. The cortical surface was sampled, yielding a mesh of 10002 grid points. Principal component analysis was performed on estimated sources (a matrix of size $201\times 10002$), yielding a strong contribution of the first principal component (PC1, 56\% of inertia, to be compared with 12\% for the second PC). Corresponding time loadings are displayed in \textbf{Fig.~\ref{fi:profil}}, and form the time profile that was actually used in our simulations. Space (i.e. cortical) loadings are displayed in \textbf{Fig.~\ref{fi:sources}}, top left. The role of parcels appears clearly, as well as the spatial sparsity of the estimated sources (which is a by-product of the gaussian mixture model). Projections onto the cortical surface of PC1 show a strong localization in the frontal area, which is the expected localization for slow waves.


\section{Conclusion}
\label{sec:conclusion}
We have presented in this paper first results on an extended wavelet MEM algorithm for time-space source localization from MEG measurements. The goal of this extension is to account explicitely for time correlations in the source space, which are not exploited (or exploited implicitely by a change of representation space, i.e. going to wavelet or time frequency space) in most classical approaches.
The resulting curse of dimensionality is addressed using various dimension reduction tools.

As a result, our numerical simulations (from realistic data) confirm that accounting for time correlations indeed improves precision in terms of time resolution, and that the sparsity properties induced by the gaussian mixture reference distribution also yields significant improvements in terms of detection performances.

A main difficulty of the approach lies in the choice of the model parameters. In this paper, some of these parameters were estimated from a first quick inversion, while some others were chosen by the user. Fully adaptive parameter choice would be desirable. Also, a natural follow up would be the study of spatio-temporal networks an the source level, for example exploiting space-time source covariance matrices.

These questions will be addressed in a forthcoming publication, together with a more complete simulation study and extensive applications to real data.

\subsection*{Acknowledgements}
Part of this work was done while M.C. Roubaud and B. Torr\'esani were visiting the Centre de Recherches Math\'ema\-tiques (CRM, UMI 3457) at Universit\'e de Montreal, both wish to thank CNRS for support and CRM for hospitality.



\bibliographystyle{IEEEbib}
\bibliography{KMEM}

\end{document}